\documentclass[10pt,twocolumn,twoside]{IEEEtran}
\usepackage[usenames,dvipsnames,svgnames,table]{xcolor}
\ifCLASSINFOpdf
  \usepackage[pdftex]{graphicx}
  \usepackage{enumitem}
  \usepackage{array}
  \DeclareGraphicsExtensions{.pdf,.jpeg,.png}
\else
  
\fi

\usepackage{subfig}
\usepackage{ulem}

\usepackage[cmex10]{amsmath}

\usepackage{stfloats}

\begin{document}
%
\title{Digital Image Forensics vs. Image Composition: An Indirect Arms Race}
%
%
%

\author{Victor~Schetinger,
        Massimo~Iuliani,
				Alessandro~Piva,~\IEEEmembership{Senior Member,~IEEE,}
        and~Manuel~M.~Oliveira
				}

%
%

\markboth{IEEE Transactions on Information Forensics and Security}%
{Shell \MakeLowercase{\textit{et al.}}: Bare Demo of IEEEtran.cls for Journals}
%



\maketitle

\begin{abstract}
The field of image composition is constantly trying to improve the ways in which an image can be altered and enhanced. While this is usually done in the name of aesthetics and practicality, it also provides tools that can be used to maliciously alter images. In this sense, the field of digital image forensics has to be prepared to deal with the influx of new technology, in a constant arms-race. In this paper, the current state of this arms-race is analyzed, surveying the state-of-the-art and providing means to compare both sides. A novel scale to classify image forensics assessments is proposed, and experiments are performed to test composition techniques in regards to different forensics traces. We show that even though research in forensics seems unaware of the advanced forms of image composition, it possesses the basic tools to detect it. 

\end{abstract}


\IEEEpeerreviewmaketitle

\section{Introduction}
The popularization of digital cameras and 
the Internet have made it easy for anyone to capture and share pictures. 
Current research suggests, however, that people are not very keen on discerning between real and edited pictures ~\cite{DBLP:journals/corr/SchetingerOSC15}. 
This poses a critical problem, as 
softwares such as Adobe Photoshop~\cite{Photoshop} and GIMP~\cite{gimp} allow anyone to easily create high-quality composites.
In such a scenario, how can one be sure that an image is authentic and depicts a factual scene? 

An arms race between forgers and forensics analysts is in progress~\cite{SurveyFarid}. While new and more sophisticated forges are being conceived, forensic techniques keep evolving to catch them. 
Most image manipulations, however, are neither malicious nor dangerous. There are plenty of legitimate reasons to edit images, such as for marketing and design. Unfortunately, sophisticated tools developed for these tasks can be used by forgers and the analysts have to struggle to catch up.

In this paper, we analyze the current state of this arms race between the field of image forensics and 
image composition techniques. Here, {\it image composition} is used as an umbrella term for all
techniques from areas such as computer graphics, computational photography, image processing, 
and computer vision, that could be used to modify an image. More specifically, we discuss works that have the potential to either be used to perform or hide forges in digital images.

From a forensic point of view, image manipulation is usually classified as either {\it splicing}, {\it copy-pasting} (also called {\it cloning}), {\it erasing}, or {\it retouching}. 
A splicing operation consists of transferring an object from an image into another, but this could be done simply by cutting and pasting, or using an advanced technique that matches the gradient of the target image~\cite{Poisson2003}. Copy-pasting is similar in essence, but the transferred object comes from the same image.
Retouching has a vague definition that could fit a wide range of actions, such as blurring regions of the image, recoloring, and applying filters.

While many modern image-composition techniques could be 
used to make sophisticated forgeries,
almost none of them have been scrutinized by forensic works.
%
%
%
There is, however, a large body of
forensic tools that could be used for this task.
This paper surveys both the forensics arsenal and image-composition techniques to identify the best strategies to analyze these novel forgeries.
We conclude that the field of digital image forensics is prepared to deal with image composition.

%
\begin{figure*}[ht!]
\centering
\subfloat[]{\includegraphics[width=0.2339\textwidth]{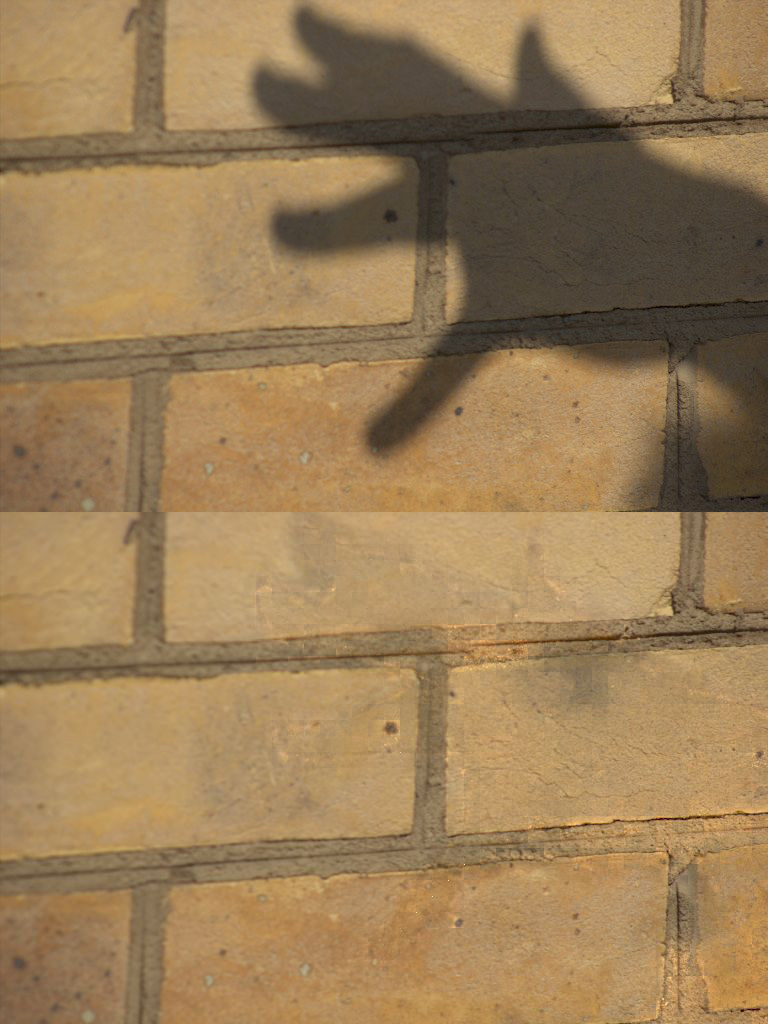}
\label{fig:unshadow}}
\subfloat[]{\includegraphics[width=0.416\textwidth]{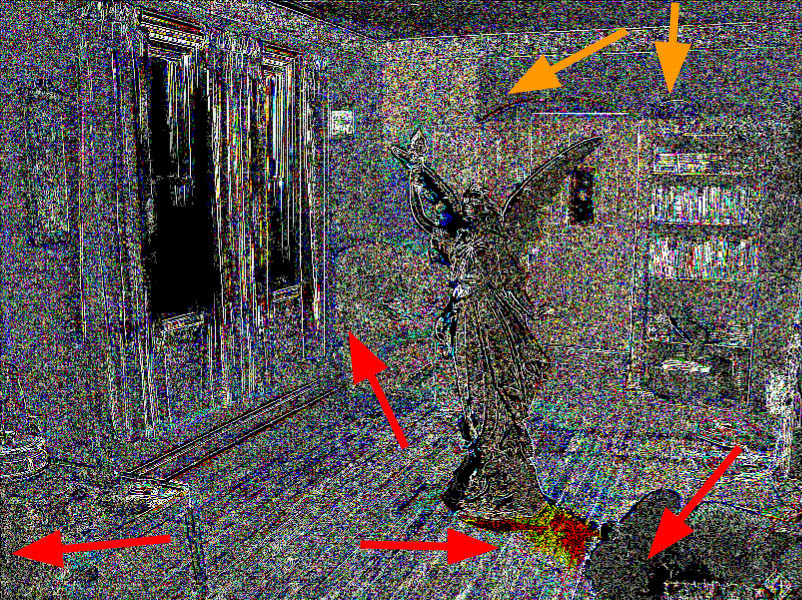}
\label{fig:insert2}}
\subfloat[]{\includegraphics[width=0.2339\textwidth]{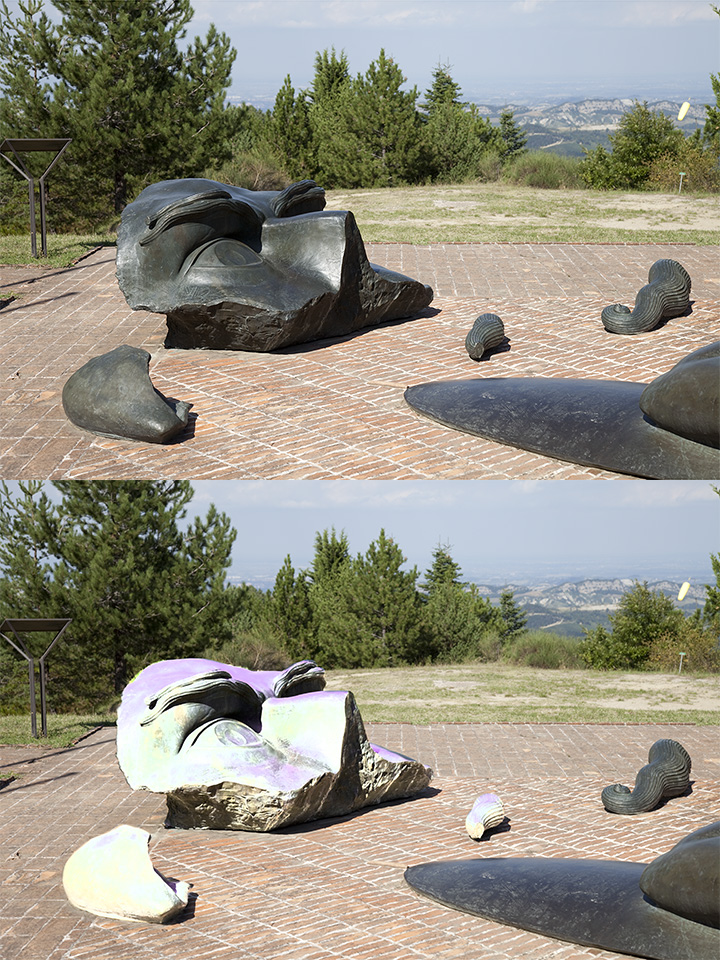}
\label{fig:localizedrecoloring}}
\\
\subfloat[]{\includegraphics[width=0.208\textwidth]{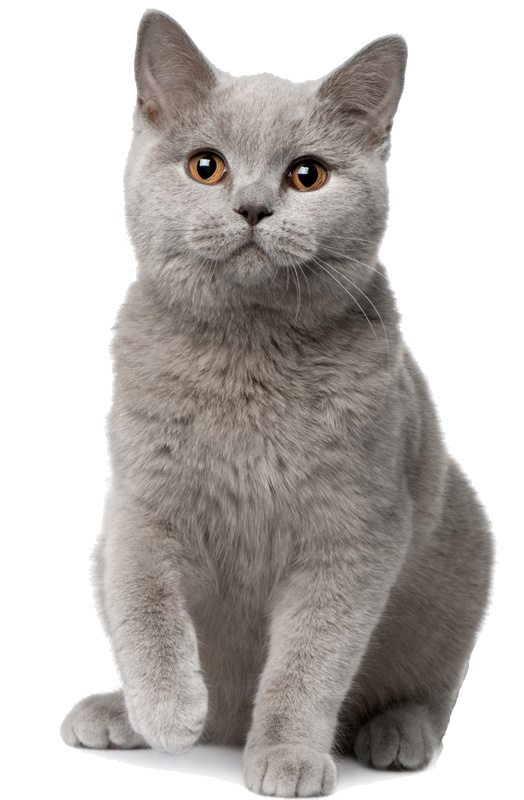}
\label{fig:morph1}}
\subfloat[]{\includegraphics[width=0.208\textwidth]{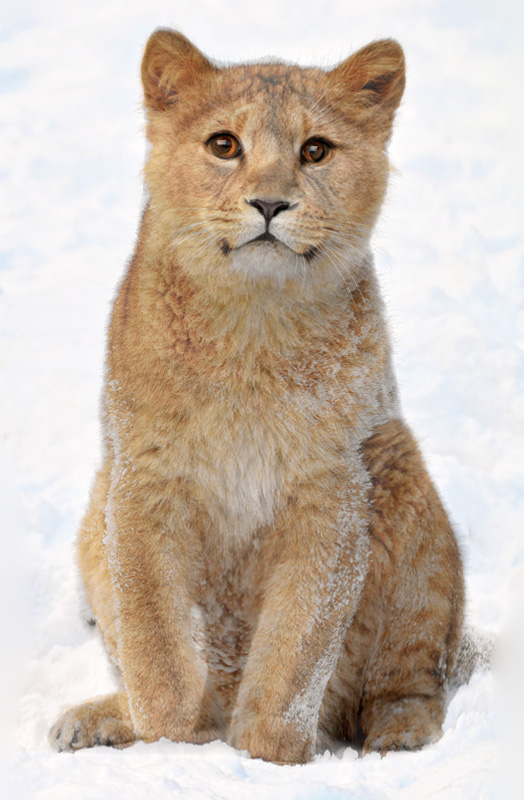}
\label{fig:morph2}}
\subfloat[]{\includegraphics[width=0.208\textwidth]{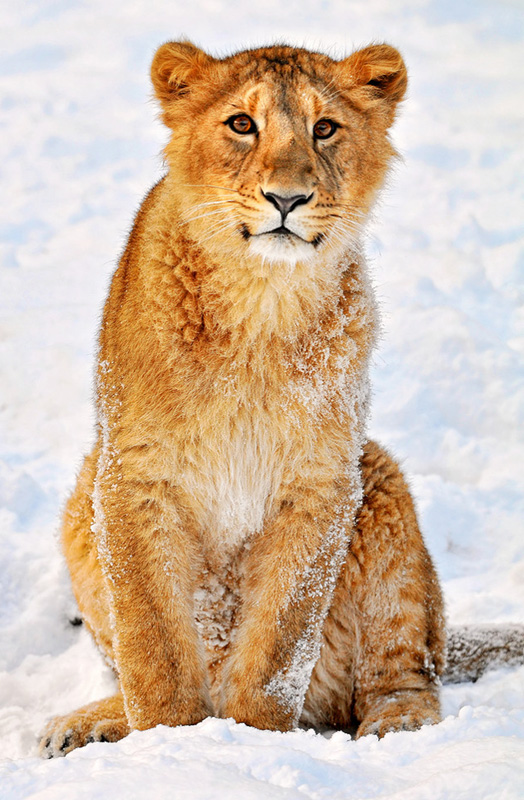}
\label{fig:morph3}}
\subfloat[]{\includegraphics[width=0.256\textwidth]{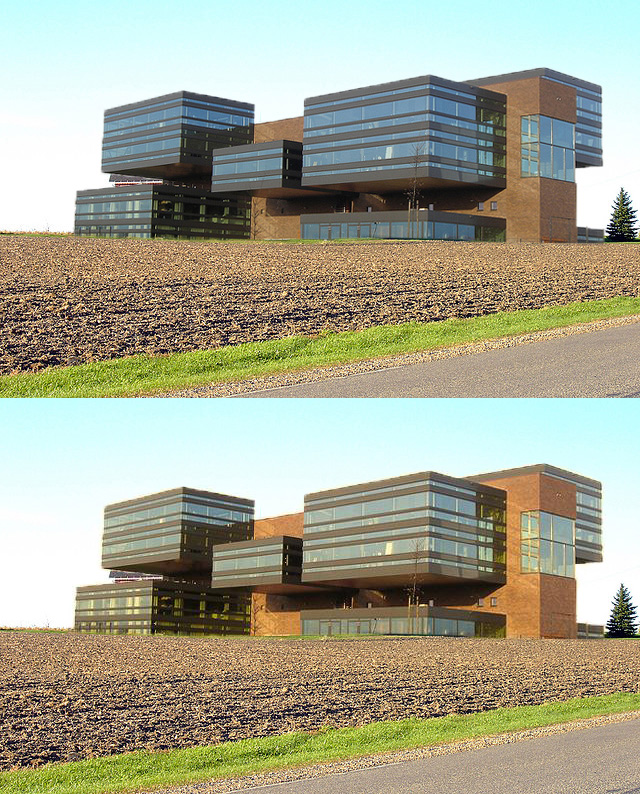}
\label{fig:reillum}}
\caption{Different ways in which composition techniques can be used to alter images. a) Removing soft shadows~\cite{Gryka2015softShadows}. The hand shadow from the top image has been removed on the bottom image. b) Inserting synthetic objects~\cite{Karsch2014}. The marble angel in the picture is not real, it was rendered into the scene along with its complex light interactions. c) Performing edge-aware filtering~\cite{GastalOliveira2015}. The bottom image was filtered to perform a localized color editing on some of the stone statues. d-f)  Morphing two different objects together to create a blend~\cite{ImageMorphing2014}. The cat in Figure~\ref{fig:morph2} is a composite of the cat in Figure~\ref{fig:morph1} and the lion in Figure~\ref{fig:morph3}. g) Transferring an object from one image to another,  adjusting its illumination according to the target scene~\cite{Xue2012}: the building was spliced on the field in the top image, and in the bottom it had its lighting adjusted to match the composition.}
\label{fig:compositingexamples1}
\end{figure*}
%
%

The main {\bf contributions} of this paper include: 
\begin{itemize}
\item A novel scale for the different levels of forensics image assessments (Section \ref{sec:FDscale}); 
\item A new way to classify forensic techniques, combining the two most used classifications in the literature with the proposed scale (Section \ref{sec:ForensicArsenal}); 
\item A new classification for image composition, extending the existing types worked in the forensics literature (Section \ref{sec:ImageComposition});
\item A theoretical and experimental analysis on image composition techniques from a forensics point of view (Sections \ref{sec:ImageComposition} and \ref{sec:TestingAndResults}). 
\end{itemize}

\section{The Forgery Detection Scale}
\label{sec:FDscale}
According to recent surveys~\cite{SurveyFarid}\cite{Rocha_2011}\cite{SurveyPiva}, there are two main fronts to digital image forensics: source identification and forgery detection. Source identification tries to link an image to the device that took it. Acquisition traces are extracted from the investigated image and then compared with a dataset of possible fingerprints specific for each class/brand/model of devices. Matching fingerprints are exploited to identify the image source device.
%
%
%
Forgery detection is concerned in determining if and how a target image has been altered. In the same sense, an image is provided for analysis, several different image features and traces can be observed, and there are different possible outcomes. This paper is focused on forgery detection, but our findings can provide useful insights for device identification.

Forgery detection mostly works twofold: either looking for patterns where there shouldn't be, or looking for the lack of patterns where there should be. If someone, for example, splices an object from an image into another, and resizes it so it is in the same scale as the target picture; the resizing operation now creates a pattern of correlation between its pixels where there shouldn't be one. On the other hand, the target image had several patterns, such as its CFA (Color Filter Array, explained on Section~\ref{sec:AcquisitionTraces}), the illuminant color, etc. which were disrupted by the spliced object.

The most simplistic view would pose that there are only two outcomes for forgery detection: the image has been altered, or no evidence of alteration is found. However, this classification might not be sufficient. Simply compressing an image might be considered an alteration, even though it is a commonplace operation, making this classification useless. Different forensic techniques work on different assumptions of what traces could be present on the image and what they can infer from them, from the location to the nature of the forgery. There is no standard on the literature, however, for classifying and comparing techniques based on their outcomes.

Here, we propose a new general classification scale called the FD (short for Forgery Detection scale). This scale is based in the concept of an image being native or not: a native image is an image that was captured by a device and then outputted to the user ``as-is". Conceptually, this is easy to define, but technically there might be some complications: different devices process the image differently. These problems will be discussed in detail further in the paper. 

The FD scale ranks forensic techniques based on the type of evidence they can provide about an image's nativity. The first possible outcome is the negative case, when it is not possible to determine if an image has undergone any form of alteration. This could happen because the image is really native, or because the analyzed traces do not show forgery, but it makes no difference: it is not possible to say that an image is truly native, only that there is no evidence supporting it to be forged. This outcome falls outside our scale in practice, but can be called FD0 for simplicity.
The following are the different levels of our Forensics Detection scale:

\begin{itemize}
\item [\textbf{FD0}] No evidence can be found that the image is not native.
\item [\textbf{FD1}] The image has undergone some form of alteration from its native state, but the nature and location of it is unknown. 
\item [\textbf{FD2}] The image has undergone some form of alteration from its native state and the location of the alteration can be determined, but its nature is not known.
\item [\textbf{FD3}] The image has undergone some form of alteration from its native state, the location of the alteration can be determined and the nature of it is known.
\item [\textbf{FD4}] All the conclusions of the previous item, and a particular tool or technique can be linked to the forgery. 
\end{itemize}

These should be referred as FD (short for forgery detection) scale, with values FD0-4. It could be argued that the ultimate form forgery detection would go beyond identifying the used technique, by locating a forger or even estimating the historic of the image's alterations~\cite{dias2012image}. It is a valid point, but they are not common in digital image forensics and fall out of this paper's scope. The FD scale is backwards inclusive for FD $> 0$, meaning that if FD4 can be guaranteed so can FD3, FD2 and FD1.
The following subsections provide in-depth explanation of the different levels of the FD scale and important considerations.

\subsection{FD1: Nativity}
The FD1 level differs from the negative case FD0 because it is possible to determine that the image is not native. This is not so simple to assess, as most modern cameras have a processing pipeline comprised of several operations (demosaicing, white balance, etc), changing the image before the user has access to it. Furthermore, demosaicing is such a fundamental operation in modern cameras that it makes little sense talking about images without it.
For the sake of generality, we propose that any form of pre-processing on the image up until a single compression can be accepted without breaching the image nativity. In the cases where the capture device is known, a forensics specialist can limit the scope of his analysis and determine only some operations are valid.
A forensic technique achieves FD1 when it is able to find evidence of alteration after capture. Techniques that analyze an image's EXIF information are an example of FD1: they can detect an inconsistency in the metadata proving an image is not native, but nothing can be said about location or nature of the alteration.

\subsection{FD2: Location}
The FD2 level is obtained when the general location of the alteration in the image is known. 
It is possible that a region of an image has been erased by a series of copy-pasting operations, and then retouched with smoothing brushes. In this case the boundaries of the forgery might no be as clear. If a technique is able to obtain any form of specificity on the altered region, FD2 is achieved. This is the case when analyzing traces such as PRNU (Photo Response Non-Uniformity, Section~\ref{sec:AcquisitionTraces}), CFA or ELA (Error Level Analysis, Section~\ref{sec:CodingTraces}), that are locally structured on the image.
If evidence of any global alteration on the image is found, such as median or bilateral filtering, then the location of the forgery is the whole image. Similarly, operations that remove parts of the image such as seam carving and cropping can be detected but the actual altered area is not present in the analyzed image anymore. It is argued that the forgery location can be considered to be all image, reaching FD2.

\subsection{FD3: Nature}
The nature of the forgery can be subjective, because it is not possible to predict all ways in which an image can be altered. The most commonly studied forms of forgery such as splicing, copy-pasting and erasing, are just a subset of possibilities. As was discussed on the introduction, image composition techniques are able to alter the shape, texture and orientation of objects, and even merge them together. 
For simplicity, any meaningful information in addition to location of the processing that can be used to assist the forensics analyst can be considered FD3. If an object has been rotated and scaled, identifying any of these operations awards an FD3 level on the scale. Identifying a spliced object is worth an FD3 on the scale because the image is not native (FD1), its location on the target image is evident (FD2), and the nature of the alteration is known (FD3).

\subsection{FD4: Technique}

The highest level on our scale, FD4, is achieved when the analyst finds evidences that can link the forgery to a particular technique or tool. A splicing can be done by simply cutting a region from an image an pasting over another, but there are also sophisticated ways to blend them, such as Alpha Matting or Seamless Cloning. A forensic technique that is able to, after obtaining FD3, provide further insight into the technique or tool used to perform the forgery achieves FD4.


\section{The forensics arsenal}
\label{sec:ForensicArsenal}
The current state-of-the-art on digital image forensics provides an arsenal of tools and techniques for forensics analysts. 
In this section we investigate the most relevant approaches and their capabilities, both in terms of applicability (i.e. when we can use them) and assessment (i.e. the level that can be achieved in the FD scale). In a general way, it can be noted that there is a trade off between the generality and the FD level that a technique is able to reach. This is intuitive, because the higher the level on the scale, the more specific the assessments are. FD1 can be simplified as a boolean statement (the image is either native or not). From FD2 onwards, there is a large set of possible answers (all different combinations of pixels in the image). To identify the nature of the forgery in the image (FD3), a technique must be looking for more specific features or traces. 

An image forensic tool is usually designed considering three steps:
\begin{enumerate}
	\item Some \textbf{traces} in the image - possibly introduced by the forgery process - are identified; such traces can be scene-level information such as ``the image lighting", or signal-level, such as ``the color filter array pattern".
	\item These traces are measured and quantified in some way, resulting in \textbf{features}, which are usually numeric in nature; 
	\item By analyzing through experimentation how the set of features behaves on native and forged images, a \textbf{decision} is taken about the image. This can be done using simple thresholds or sophisticated machine learning techniques.
\end{enumerate} 
%
%
%

\begin{table}[ht!]
\centering
\caption{The steps of the tool by Carvalho et. al.~\cite{Tiago2013}.}
	\begin{center}
		\begin{tabular}{||c|c||} 
			\hline
			\textbf{Layer} & \textbf{Example} \\ 
			\hline
			Trace    & Illuminant or light \\ 
			&	source of the image.     \\
			\hline
			Feature    & Estimated illuminant colors \\
			& and light intensity on object edges. \\
			\hline
			Decision & SVM classification.             \\
			\hline
		\end{tabular}
		\label{table:example}
	\end{center}
\end{table}

In Table \ref{table:example} we show a practical example of previous steps for the technique developed by Carvalho et. al.~\cite{Tiago2013} to detect splicing. The used \textbf{trace} is the illuminant, or the light source. The key observation is that if an object is spliced and the original image had different light conditions, such as indoor or outdoor lighting, or even incandescent vs. fluorescent lights, this trace can be used to identify it. The \textbf{features} used are the estimated illuminant colors and the light intensity on the edges, for the different analyzed regions of the image. The \textbf{decision} process uses a Support Vector Machine (SVM) to classify the image as either spliced (FD3) or inconclusive (FD0) based on the features.

%
%

\begin{figure}[]
\includegraphics[width=0.49\textwidth]{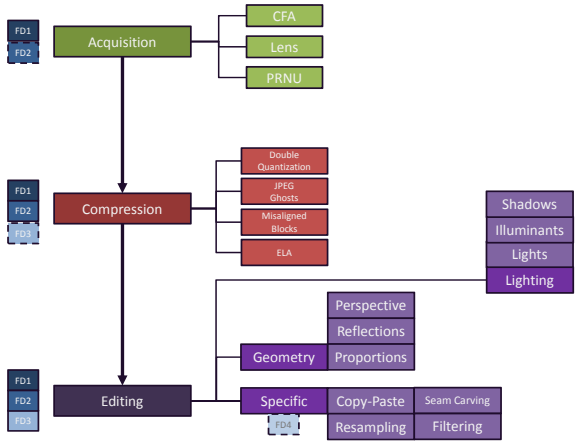}  
\caption{Forensic techniques' classification. Each type of trace is organized under its correspondent phase on the forgery process. The techniques themselves were omitted for the sake of clarity, but would appear as leaf nodes under their analyzed traces. On the left, the relation to the FD scale is displayed. Only by analyzing specific editing traces it would be possible to achieve FD4.}
\label{fig:classification}
\end{figure}

Forensic tools classification is based on the traces they analyze. Piva~\cite{SurveyPiva} distinguishes between traces left by three different steps of the image formation process: acquisition, coding and editing. Another intuitive classification has been proposed by Farid~\cite{SurveyFarid} where the forensic techniques are grouped into five main categories: pixel-based, format-based, camera-based, physically-based and geometric-based. Farid's classification is more common in the literature and distinguishes better between traces, but Piva's can be closely related to the FD scale. We propose a classification based on Piva's approach (Fig.~\ref{fig:classification}), but with greater specificity, similarly to Farid's.

%
 %


The most relevant traces and correspondent tools developed by the forensic community will be discussed in the following subsection. The FD scale will be used to describe which level of assessment can be expected when examining an image using a specific tool.


\subsection{Acquisition Traces (AT)}
\label{sec:AcquisitionTraces}
Native images come to life with distinctive marks (artifacts, noise, inconsistencies) due to the acquisition process. 
Both hardware (e.g., lens, sensor) and software components (e.g., demosaicing algorithm, gamma correction) contribute to the image formation, introducing specific traces into the output (native) image.
When a native image is processed some of these traces can be deteriorated or destroyed, exposing evidence of tampering. Forgery detection using acquisition traces generally falls in one of two categories:
		\begin{enumerate}
		\item \emph{Global}: The analyzed trace is a global camera signature. Non-native images can be exposed when this signature does match with the supposed source device. 
		For instance, in~\cite{farid-JPEGheader} non-native JPEG images are exposed by analyzing quantization tables, thumbnails and information embedded in EXIF metadata.
		In~\cite{LukasImageOrigin} the reference pattern noise of the source device is taken as a unique identification fingerprint. The absence of the supposed pattern is used as evidence that the image is non-native.
		\item \emph{Local}: The analyzed trace has a local structure in the image. Its inconsistencies in some portion of the image can be exploited to localize the tampering.
		For instance, Ferrara~\cite{ferrara} uses demosaicking artifacts that form due to color interpolation. They can be analyzed at a local level to derive the tampering probability 
		of each 2$\times$2 image block. Fridrich~\cite{FridrichPRNU} reveals forgeries by detecting the absence of the PRNU on specific regions of the investigates image.
		\end{enumerate}
		
Let us note that some traces can be considered both at a global or local level (e.g., demosaicing artefacts and PRNU), allowing to identify non-native images (FD1) or to localize forgeries (FD2). The analysis of acquisition traces is usually limited to matching a known pattern, and they can be easily disrupted. For this reason, FD2 is the highest we can expect to achieve on the FD scale using acquisition traces.
The analysis of acquisition traces generally requires some additional information about the source device. In some cases this information depends on the source device model or manufacturer (e.g., color filter array pattern, quantization tables), and can be easily obtained to assess image nativity ~\cite{JPEGsnoop}. In other cases these traces are unique camera fingerprints (e.g. PRNU) and can be obtained by having the source device available, or be estimated by using different images captured by the same device.



\subsection{Coding Traces}
\label{sec:CodingTraces}
	Lossy compression might happen in many occasions during the life of a digital image:
	\begin{itemize}
		\item Native images of non professional cameras and smartphones usually come to life in JPEG format;
		\item when uploading a photo on a social network lossy compression is usually applied to the image;
		\item when a JPEG image is altered and saved again in JPEG, double lossy compression occurs. 
	\end{itemize}
	For this reason, most of the literature has focused on studying the traces left by single and multiple chains of JPEG-compression. This is a very prolific field of study in forensics, with a wide variety of techniques. 
	Fan~\cite{FanJPEG} and Luo~\cite{LuoJPEG} provide efficient methods to determine whether an image has been previously JPEG compressed, and, if so, are able to estimate some of the compression parameters. Further advances have been also provided by Li et al.~\cite{Li15:HighQJpeg} to identify high-quality compressed images basing on the analysis of noises in multiple-cycle JPEG compression.
	On Bianchi's technique~\cite{BianchiJPEG}, original and forged regions are discriminated in double compressed images, either aligned (A-DJPG) 
	or nonaligned (NA-DJPG) even when no suspect region is detected.
	 Yang et al.~\cite{Yang14:DJPEG} propose an error-based statistical feature extraction scheme to face the challenging case where both compressions are based on the same quantization matrix.
	
	In most cases the analyst can exploit coding traces to disclose non-native images or to localize the tampering, 
	reaching FD1 and FD2 in the forensic scale;
	FD3 has not been deeply investigated but, as shown in literature, coding traces can reveal something more
	than mere localization of the tamper.
	Farid~\cite{Farid:Ghost} shows that, when combining two images with different JPEG compression quality, 
	it may be possible to recover information of the original compression quality of the tampered region. 
	A stronger compression in later stages usually deteriorates the traces of previous compressions, compromising the effectiveness of these techniques.
	This technique has been proved effective only if the tamper was initially compressed at a lower quality 
	than the rest of the image; on the contrary, when the compression is stronger in the latter stage, 
	the traces of the first compression are probably damaged and the detection fail. 
	
	This behavior is common in most forensic techniques based on the analysis of multiple compression chains: 
	stronger compression in later stages ruins the traces of previous compression stages, making the tools ineffective. The analyst, when using these tools, should take into account the reliability of the results based on the coding characteristics of the investigated image~\cite{Ferrara2015}.


\subsection{Editing Traces}
	Image editing modifies the visual information of the image and the scene depicted, introducing traces in several domains of the image such as pixel, geometric, and physical. Editing traces are the most numerous, and can be split into subcategories (Fig.~\ref{fig:classification}) according to these domains. 
	
	Image illumination inconsistencies (light source direction, cast and attached shadows) are powerful traces considering that it is hard to achieve a perfect illumination match when composing two images. The are two main approaches for illumination techniques: geometric and illuminant. The first are based on the geometric constraints of light, trying use scene elements as cues to determine if the arrangement of lights~\cite{FaridLightingEnvironments}\cite{Saboia2011}\cite{Kee2009}\cite{FaridComplex}\cite{Carvalho2015} or shadows~\cite{Kee:2013}\cite{Kee2014} are plausible. Illuminant techniques exploit the color, intensity and temperature aspects of the illumination, and are able to detect if a region or object in the image was lighted by a different type of light~\cite{Riess2010}\cite{Carvalho:Color}.
	
	Similarly, geometric relations within an image (e.g., object proportions, reflections) are determined by the projection of the 3D real scene onto the image plane. This process is commonly modelled through the pin hole camera model~\cite{Hartley2004}. Any deviation from this model can be exploited as evidence of tampering. 
	Yao~\cite{Yao:Perspective} uses a perspective constrained method to compare the height ratio between two objects in an image. Without the knowledge of any prior camera parameter, it is possible to estimate the relative height of objects and eventually identify  objects that have been inserted on the scene. An extension has been proposed by Iuliani et al.~\cite{Massimo2015} to apply the technique on images captured under general perspective conditions.
	Conotter~\cite{Conotter:2010:sign} describes a technique for detecting if a text on a sign or billboard has been digitally inserted on the image. The method looks if the text shape satisfies the expected geometric distortion due to the perspective projection of a planar surface. The authors show that, when the text is manipulated, it is unlikely to precisely satisfy this geometric mapping.
	
When an editing trace exposes evidence of forgery, we can expect to infer something about its nature (FD3): if an object has as shadow inconsistent with the scene, he was probably inserted; if the illuminant color is inconsistent, the object could have been either spliced or retouched. 

Obtaining other specific information about the techniques involved in the tampering process (FD4) is a very challenging task. There are two main reasons for this. The development of a technique for detecting the use of a specific tampering process/tool may require a strong effort compared to its applicability in a narrow range. Secondly, proprietary algorithms have undisclosed details about their implementation, making hard to develop analytical models for their traces. A first step toward this kind of assessment have been proposed by Zheng et al.~\cite{zheng2015exposing} to identify the feather operation used to smooth the boundary of pasted objects.


\section{Image Composition}
\label{sec:ImageComposition}
The term ``Image Composition" is used to encompass different fields such as Computational Photography, Image Processing, Image Synthesis, Computer Graphics and even Computer Vision. Recent works on all of these fields were surveyed to determine which ones could be used to aid in forgery. For this purpose, techniques that a forger could use to perform any form of operation were considered, from splicing to highly creative operations.

The techniques were classified in five general classes based on the type of forgery they could perform:
\begin{itemize}
\item \textbf{Object Transfering}: transfering an object or region from one image to another image, or even to the same image. This is the most common type of forgery, and encompasses both splicing and copy-and-paste operations. It is divided into \textit{Alpha Matting}, \textit{Cut-Out}, \textit{Gradient Domain}, \textit{Structurally Changing}, \textit{Inpainting}, and \textit{Composition}; 
\item \textbf{Object Insertion and Manipulation}: inserting synthetic objects into an image or manipulating an existing object to change its properties. It is divided into \textit{Object Insertion}, \textit{Object Manipulation}, and \textit{Hair};
\item \textbf{Lighting}: altering image aspects related to lights and lighting. It is divided into \textit{Global Reillumination}, \textit{Object Reillumination}, \textit{Intrinsic Images}, \textit{Reflections}, \textit{Shadows}, and \textit {Lens Flare};
\item \textbf{Erasing}: removing an object or region from the image and concealing it. It is divided into \textit{Image Retargeting}, and \textit{Inpainting};
\item \textbf{Image Enhancement and Tweaking}: this is the most general class of forgery, and is related to what is considered retouching on the forensics literature. It is divided into \textit{Filtering}, \textit{Image Morphing}, \textit{Style Transfer}, \textit{Recoloring}, \textit{Perspective Manipulation}, and {Restoration/Retouching}.
\end{itemize}

It must be noted that this classification considers forgery from the point of view of both Image Composition and Forensics. A composition technique that can be used to transfer an object from one image to the other can also transfer it to the same image. While those are distinct forgeries (and problems) from the forensics point of view, they can be performed by the same technique. Furthermore, some of the surveyed techniques could be used to perform more than one type of forgery in the classification. Erasing, for instance, is often performed by copy-pasting regions of the image to conceal an object. In this sense, a technique under the \textit{Object Transfering} classification can be also considered on the \textit{Erasing} class.


In this Section, we discuss each of the different forgery classes and their relation to the forensic traces and techniques. Most information about the effect of composition on forensic traces comes from performed tests (see Section~\ref{sec:TestingAndResults}). 

\subsection{Object Transferring}

This class contains techniques that can be used with an end goal of transferring objects between images or in the same image. A fundamental task of transferring an object or region is defining its boundaries, and techniques that can help on making good contours are classified as \textit{Cut-out}~\cite{Mortensen1995}\cite{HuangRepSnapping}. These techniques do not change the content from the source or target images, they only aid in selecting a pixel area. 

Most techniques to detect splicing or copy-and-paste are well-suited against \textit{Cut-out} forgeries, because the pixel content is unaltered. From a forensics point of view, well-defined boundaries on the transferred region reduce the amount of information being carried from the original image. This might alter some traces and affect the performance of techniques based on those traces~\cite{Sutthiwan2010}. A bad cut can also be easy to note visually, without the use of additional tools. 

One of the main limitation of transferring objects by cut-and-paste is that transparency is ignored. Hair, thin fabrics, glass, and edges may contain a mix of colors from the foreground and background of the source image. This can cause visual artifacts on the resulting composition, and the presence of foreign colors that can be used for traces. \textit{Alpha Matting} techniques can estimate the transparency of a region in the image, which can be used to better extract it from the source image, and then composite on the target image (Fig. 3e-h).
The visual aspect is the most critical on the use of alpha matting for object transferring, as it blends colors on borders and transparent regions, making convincing forgeries. In our tests, there was almost no difference between detecting Alpha Matting or regular cut-out.

The most sophisticated object transferring techniques are \textit{Gradient Domain} ones. These techniques aim to combine the gradient of the transferred object with the target image, making a complex blend. The simplest technique is Poisson Image Editing~\cite{Poisson2003}, which matches the gradients by solving a Poisson equation from the boundaries of the transferred region. The resulting object has different colors and gradient, blending with the scene. Poisson Image Editing, also commonly referred to as \textit{Seamless Cloning}, spawned several works that improved on its basic idea of solving differential equations for gradient matching on transferred regions~\cite{ddp_siggraph2006}\cite{Yang2009}\cite{Farbman2009}\cite{Tao2010}\cite{Ding2010}.

\begin{figure*}[ht!]
\centering

\subfloat[]{\includegraphics[width=0.24\textwidth]{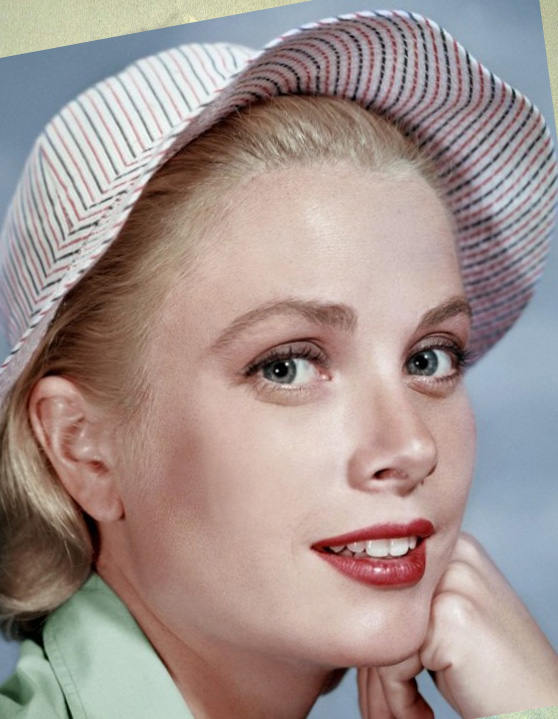}
\label{fig:harm1}}
\subfloat[]{\includegraphics[width=0.24\textwidth]{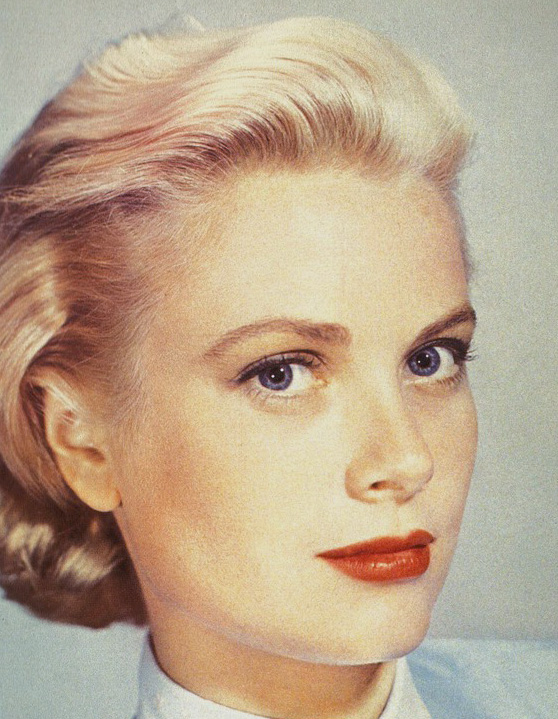}
\label{fig:harm2}}
\subfloat[]{\includegraphics[width=0.24\textwidth]{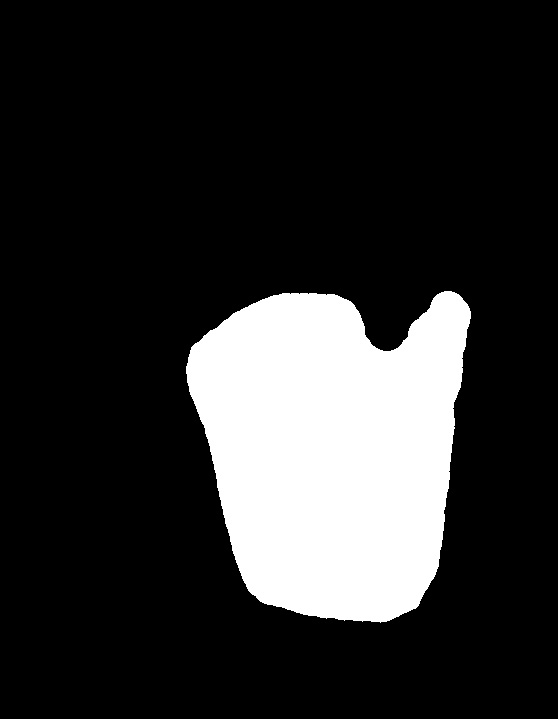}
\label{fig:harm3}}
\subfloat[]{\includegraphics[width=0.24\textwidth]{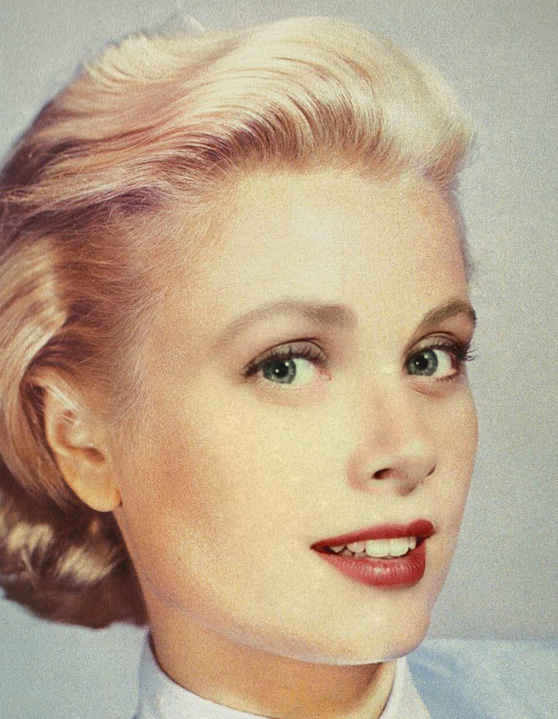}
\label{fig:harm4}}
\\
\subfloat[]{\includegraphics[width=0.24\textwidth]{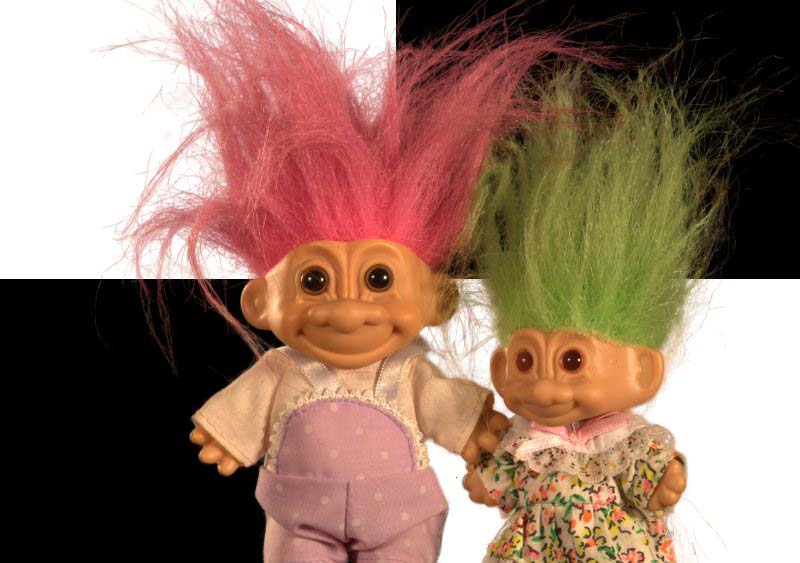}
\label{fig:am1}}
\subfloat[]{\includegraphics[width=0.24\textwidth]{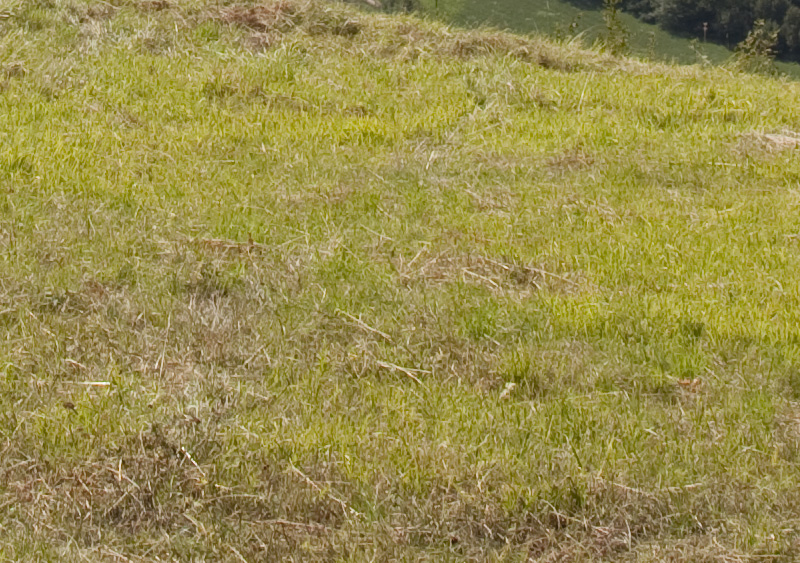}
\label{fig:am2}}
\subfloat[]{\includegraphics[width=0.24\textwidth]{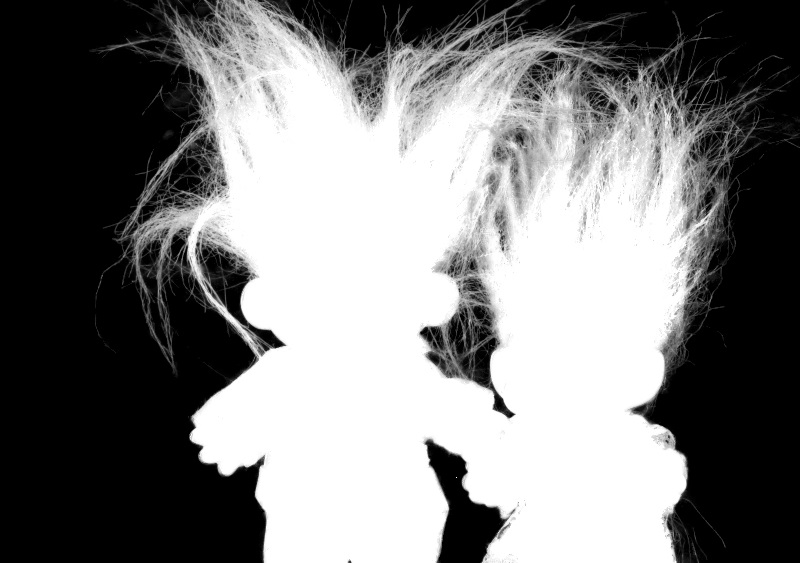}
\label{fig:am3}}
\subfloat[]{\includegraphics[width=0.24\textwidth]{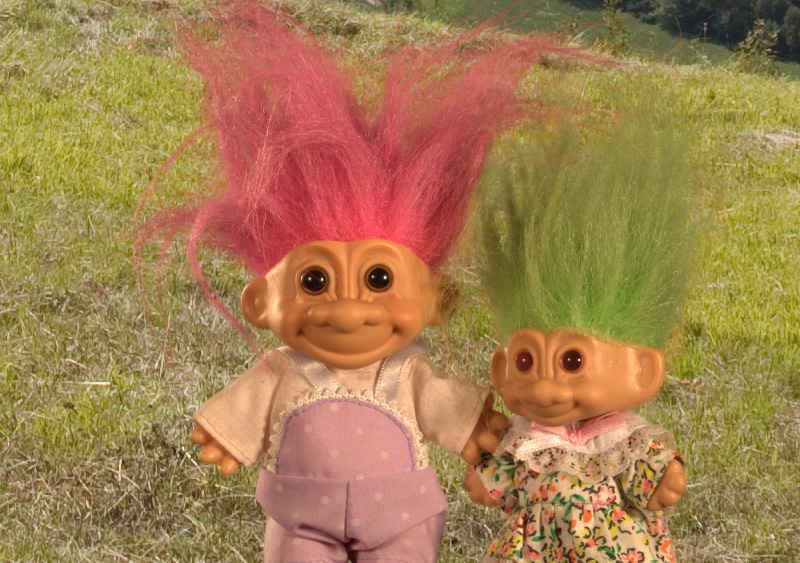}
\label{fig:am4}}
\caption{Example of splicing using object transferring techniques. The top row represents \textit{Alpha Matting}, and uses the Shared Matting technique~\cite{Gastal2010}. The bottom row corresponds to \textit{Gradient Domain}, and uses Multi Scale Harmonization~\cite{Sunkavalli2010}. The source images are on the first column, the target images on the second colum, the transference masks are on the third column, and the final result is displayed on the fourth column for each technique.}
\label{fig:compositingexamples2}
\end{figure*}

Most of the Gradient Domain techniques use Laplacian Pyramids to manipulate the different image frequencies. Works such as Sunkavalli's~\cite{Sunkavalli2010}, however, focus on the Laplacian Pyramid as the main component for sophisticated blends between images, being able to maintain the noise and texture of the target image (Figures~\ref{fig:am1} through~\ref{fig:am4}) to some degree. This kind of approach was generalized~\cite{Farbman2011CP} and improved~\cite{Darabi12} by other authors.

Gradient Domain techniques can be the most ``dangerous" for object transferring, from a forensics point of view. They can blend the whole transferred area and merge the images on a profound level. There are big variations on the inner workings of each technique, and the results are very dependent on the images to be combined. This means that it is hard to evaluate them from a forensics point of view.  A more specific and thorough study is required to determine the exact effect of compositing images on the Gradient Domain.
In our tests they were easily identifiable using compression and noise traces, but these techniques have the potential to hide such traces if finely tuned. The safest way to detect forgeries of this kind would be focusing on high-level traces such as shadows and geometry. Light-based traces could help on cases where a full object is being transferred, because the resulting colors after the blending may create irregular lighting. When transferring parts of objects, such as changing faces on an existing head (Figure~\ref{fig:harm4}), it is possible that the result can have plausible lighting and illuminant traces.


Object Transferring techniques are arguably the most relevant to the forensics community, because they can be used to perform both splicing and copy-pasting. Figure~\ref{fig:compositingexamples2} shows an \textit{Alpha Matting} (top row), and a \textit{Gradient Domain} (bottom row) splicing. Both forgeries are visually unnoticeable. Notice how the \textit{alpha matte} (Figure~\ref{fig:am3}) contains very precise information about the transparency of each hair, and the mixture of colors on the final composition (Figure~\ref{fig:am4}). The \textit{Gradient Domain} composition exemplified does not use transparency information (Figure~\ref{fig:harm3}), but is able to transfer some of the color and texture of the target image (Figure~\ref{fig:harm2}) into the transferred region of the source region (Figure~\ref{fig:harm1}). The final result (Figure~\ref{fig:harm4}) is a very convincing composition. 

\subsection{Object Insertion and Manipulation}

Images are 2D projections of a 3D scene, with complex interactions of light and geometry. To insert a new object into the image, or to manipulate existing objects, the properties of the 3D scene must be known. This is a very challenging task. Techniques under this category focus on estimating characteristics of the 3D scene or its objects, providing means to alter them on a visually convincing way.

Rendering a synthetic object into an image is a simple task if the scene lighting and camera parameters are known. Additional knowledge about scene geometry also helps to increase realism. For this reason, this is the focus of most object insertion techniques. The challenge is to obtain this information from a single image. The most advanced techniques for object insertion, developed by Karsch, are able to estimate perspective, scene geometry, light sources and even occlusion between objects. In~\cite{Karsch2011} heavy user input was needed to aid the parameter estimation, whereas in a second work (Figure~\ref{fig:insert2})~\cite{Karsch2014} most input tasks were replaced with computer vision techniques to infer scene parameters.

The manipulation of objects in an image suffers from similar problems than insertion. Scene lighting, camera parameters and geometry are required for a visually convincing composition, and the geometry of the object being modified must be also known. A slight advantage in relation to rendering synthetic objects is that the photographic texture of the modified object can be used, providing a more photo-realistic touch. It is possible to perform resizing operations on objects without directly dealing with its 3D geometry~\cite{wu-2010-resizing}, but most techniques will focus on modeling it. 

The easiest way to work with the geometry of objects in an image is to limit the scope to simple primitives. Zheng~\cite{zcczhm_interactiveImages_sigg12} focus on cube-like objects, modeling them through ``cuboid proxies", which allow for interactive transformations such as scale, rotation, and translation in real time. Chen's work~\cite{Chen:2013:EEO:2508363.2508378} uses user input to model an objects geometry through swipe operations. This technique works specially well on objects that have some kind of symmetry, such as a candelabrum or a vase, and allows changes in the geometry itself. Another solution for dealing with object geometry is to use a database of 3D models, and find one that fits with the object depicted in the image~\cite{ObjectManipulation2014}. 

Manipulating human body parts in images is a specially hard task, because human bodies vary greatly in shape, and clothes affect the geometry. This type of manipulation, however, is of special interest due to its applications in marketing photography and modeling. Zhou~\cite{Zhou:SIGGRAPH2010} uses a parametric model of the human body, and fits a photography to a warped 3D model, achieving a correspondence between body parts in the image and 3D geometry. This allows the reshaping of body parts, making a person in a picture look thinner, stronger, taller, etc. Hair manipulation is also a hot topic in image composition, with a special focus on changing hair styles after the picture has been taken~\cite{Chai:2012:SHM:2185520.2185612}\cite{Chai:2013:DHM:2461912.2461990}\cite{Weng2013}. 

Even though state-of-the-art techniques in image insertion and manipulation can create visually convincing results, they should not pose a problem for modern forensic techniques. Distinguishing between real and synthetic images is a very debated topic~\cite{Ng2013}\cite{FaridFaces}\cite{Dang-Nguyen2014}, and there are forensic techniques that focus on identifying them~\cite{Peng2014}\cite{Peng2013a}. 

The weak point for this category of image composition is in the acquisition traces. The process of rendering a synthetic object is different from capturing it from a camera, so the acquisition traces should point to the manipulation, providing FD1 or FD2 results. Similarly, when performing transformations on an object (scaling, rotating, deforming, etc.), its pixels have to be resampled, changing the acquisition traces. Resampling detection also could be used to obtain an FD3 result in these cases, while compression-based techniques could identify this type of manipulation if the original image was compressed. Kee has demonstrated that object insertion might be able to fool geometry-based lighting techniques~\cite{Kee2014}, which could also extend to object manipulation. The reason for this is that the same lighting parameters estimated to verify the integrity of the scene were used to generate the composition.

\subsection{Erasing}

An erasing manipulation is when an element of the image is intentionally removed or hidden, and not a consequence of other editing. This category is comprised mostly of Inpanting and Image Retargeting techniques.

Inpainting techniques are used to complete a region in an image, filling it with appropriate content~\cite{Bertalmio:2000:II:344779.344972}. By selecting a region that one wants erased as the region to be completed, inpainting can make objects disappear. Several works on inpainting are focused on stitching different parts of images together~\cite{Telereg2013}\cite{Kopf:2012:QPI:2366145.2366150}, or filling large gaps~\cite{DTL:SIGGRAPH-ASIA:13}. There are implementations of inpainting techniques already available on commercial editing sofware, such as Photoshop's Spot Healing Brush and Content Aware Fill tools. The main limitation of inpainting is filling regions with high amount of details, or using image features which are not local in the filling. Huang's~\cite{Huang-TOG-2014} work is capable of identifying global planar structures in the image, and uses ``mid-level structural cues" to help the composition process.

Image retargeting is a form of content-aware image resizing. It allows to rescale some elements in an image and not others, by carving seams in the image, i.e. removing non-aligned lines or columns of pixels~\cite{Avidan:2007:SCC:1276377.1276390}. The seams usually follow an energy minimization, removing regions of ``low-energy" from the image. The objects and regions that have seams removed will shrink, while the rest of the image will be preserved. This can be used to remove regions of the image by forcing the seams to pass through certain places instead of strictly following the energy minimization. Most research on image retargeting focus on better identifying regions in the image to be preserved, and choosing the optimal seam paths~\cite{AAIR:Panozzo:2012}\cite{COMPAESTH:COMPAESTH10:001-008}.

Erasing manipulations should behave in a similar fashion to object insertion and manipulation, as the modified region will not come from a photograph, but from an estimation. This affects acquisition and compression traces, provided the original images were compressed. Image retargeting has already been analyzed from the point of view of image anonymization~\cite{Dirik2014a}, and there is even a specific technique for its detection~\cite{Yin2015}. Detecting that a seam carving has been done in an image would constitute and FD4 in our scale.

\subsection{Lighting}

Lighting techniques are capable of changing the lighting of scenes~\cite{Wanat:2014:SCC:2601097.2601150} and objects~\cite{LBPDD12}\cite{Bell:2014:IIW:2601097.2601206}, inserting light effects such as reflections~\cite{Endo:2012}\cite{Sinha2012}, lens flare~\cite{Hullin:2011:PRL:2010324.1965003}\cite{Lee:2013:PRL:2600890.2600892}, and even manipulating shadows~\cite{Mohan:2007:ESS:1251554.1251618}\cite{RuiqiGuo:2011:SSD:2191740.2191929}\cite{Finlayson02removingshadows}. From a forensics point of view, lighting techniques are dangerous due to their potential of concealing other forgeries. After splicing an object in an image, for instance, a forger could add convincing shadows and change its lighting, making it harder for both human analysts and forensic techniques to detect it. Indeed, it is a concern in image composition when the source and target lighting conditions are different, and there are works focused on correcting this issue~\cite{DBLP:journals/tog/XueADR12}\cite{LopezMoreno2010698}.

Due to the variety of lighting techniques, it is hard to make a general statement about them from a forensics point of view. As always, it seems plausible that at least an FD2 result can be achieved if compression is involved in the forgery. Techniques that add shadows or change the lighting in a visually convincing way, but do not account for all lighting parameters of the scene, could fail to deceive geometry and light-based forensics analysis. Specifically identifying light inconsistencies is an FD3 in our scale.

\subsection{Image Enhancement/Tweaking}

This is a broad classification for techniques that perform image modifications and are too specific to have their own category. Image morphing techniques~\cite{ImageMorphing2014}\cite{journals/cgf/KaufmannWSSSG13} can fuse objects together, creating a composite that is a combination of them. Style transfer techniques are able to transform an image to match the style of another image~\cite{Shih:2014:STH:2601097.2601137}, a high-level description of a style~\cite{Laffont14}, or an image collection~\cite{hacohen13}\cite{Liu:2014:AAS}. In the same vein, recoloring techniques can add or change the color of image elements~\cite{Carroll:2011:IDM:2010324.1964938}, and even simulate a different photographic process~\cite{Echevarria2013}.

Filtering techniques can be very flexible, allowing for a wide variety of effects. They can be used to remove noise or detail from images (Figure~\ref{fig:localizedrecoloring})~\cite{Gastal2012}\cite{Cho:2014:BTF:2601097.2601188}, or even to add detail~\cite{GastalOliveira2015} while preserving edges. Different filters may be designed to obtain different effects. From a forensics point of view, filtering techniques can be used to remove low-level traces. A simple median or gaussian filter is able to remove compression and CFA traces, but it is easily detectable, as it softens edges. Edge-aware filtering, however, can be used to destroy such traces preserving edges. If used in a careful way, it can remove the aforementioned traces in a visually imperceptible way.

Perspective manipulation techniques allow an user to change the geometry~\cite{Lieng:2012:Eurographics}, and perspective~\cite{Carroll:2010:IWA:1778765.1778864} of a scene, or to recapture an image from a different view point~\cite{Lee:2011:RIC}. Its uses are mostly artistic and aesthetic, but these techniques could be used to forge photographic evidence. The final type of manipulation that will be discussed is Retouching. Retouching techniques aim to perform adjusts on image properties such as white balance~\cite{Boyadzhiev:2012:UWB:2366145.2366219}\cite{Hsu:2008:LME:1360612.1360669}, focus~\cite{Tao:2013:SOO}, or several at the same time~\cite{Joshi:2010:PPE:1731047.1731050}. They can also aid in performing adjustments in several images at the same time~\cite{TIM:2012}.

\section{Testing and Results}
\label{sec:TestingAndResults}

\begin{table*}[ht!]
\centering
\caption{Results for analyzing forensic traces in the sample images for composition techniques. Each composition technique can be either Identifiable by using a certain trace, Non-Identifiable, or plausibly identifiabe. Indecisive means the traces were already compromised in the tested images, and N/A that such trace could not be tested.}
\label{tab:ImageTesting}
\begin{tabular}{|l|c|cccc|}
\hline
\textbf{Composition Technique}  & \textbf{Images Tested} & \multicolumn{1}{c|}{\textbf{CFA}} & \multicolumn{1}{c|}{\textbf{Double Jpeg}} & \multicolumn{1}{c|}{\textbf{ELA}} & \textbf{Noise}            \\ \hline
Soft Shadow Removal~\cite{Gryka2015softShadows}    & 5             & N/A                      & Indecisive                       & Non-Identifiable                & Identifiable     \\ \hline
Dehazing~\cite{Fattal2014}               & 9             & Plausible                & N/A                              & Indecisive               & Identifiable     \\ \hline
Object Insertion~\cite{Karsch2014a}       & 9             & Indecisive               & Non-Identifiable                 & Identifiable             & Identifiable     \\ \hline
Reillumination~\cite{Xue2012}         & 11            & N/A                      & Identifiable                     & Plausible                & Plausible        \\ \hline
3D Object Manipulation~\cite{Kholgade2014} & 3             & Plausible                & Indecisive                       & Non-Identifiable         & Non-Identifiable \\ \hline
Image Morphing~\cite{Liao2014}         & 12            & Plausible                & Plausible                        & Plausible                & Identifiable     \\ \hline
Alpha Matting~\cite{Gastal2010}\cite{Chuang2001}          & 9             & Identifiable             & Plausible                        & Plausible                & Non-Identifiable \\ \hline
Edge-Aware Filtering~\cite{GastalOliveira2015}   & 5             & Identifiable             & Non-Identifiable                 & Plausible                & Identifiable     \\ \hline
Seamless Cloning~\cite{Sunkavalli2010}\cite{Farbman2009}\cite{Tao2010}       & 16            & Plausible                & Identifiable                     & Plausible                & Plausible        \\ \hline
\end{tabular}
\end{table*}

To understand how Image Composition techniques affect forensics traces, we gathered images from several works and performed a forensics analysis (Table~\ref{tab:ImageTesting}). The images used were either obtained from the publication website or directly from the authors. Our main goal was to analyze the images directly before and after the techniques have been applied, to understand how the traces are affected by it.

The images obtained have varying amounts of pre-processing and compression. This influences the quality of the analysis, making it harder to evaluate all cases with the same standards. Obtaining all the implementation for the techniques and generating controlled test cases could solve this problem, but it is not feasible on an analysis of this scale.

As discussed in Section~\ref{sec:ForensicArsenal}, there are several techniques based on each trace. Each technique has a different level of success and scope. For this reason, our analysis is on a higher level. We discuss the plausibility of four different traces being used to detect a composition technique, rather than the performance of each particular forensic technique. In particular, a N/A result means there were no images for that technique that could be tested for that trace. An example of this would be compression traces when all available images were uncompressed. An Indecisive result means that the tests were inconclusive for that trace. This is probably due to some form of pre-existing processing or compression that made it impossible to evaluate the images for a particular trace. A Plausible result means that there are indications of editing or clear differences between the image before and after the composition technique was used, but it is not clearly identifiable. An Identifiable result means it is possible to identify that some form of editing has occurred using that trace. A Non-Identifiable result indicates that trace is not suitable for detecting that particular composition technique.

\begin{figure}[h!]
\centering
\subfloat[]{\includegraphics[width=0.24\textwidth]{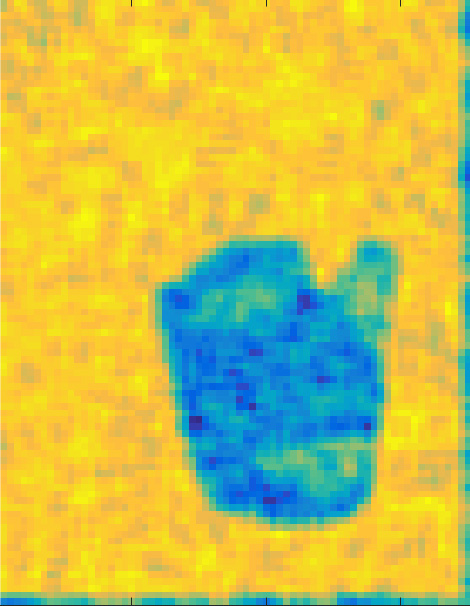}
\label{fig:f_grace}}
\subfloat[]{\includegraphics[width=0.24\textwidth]{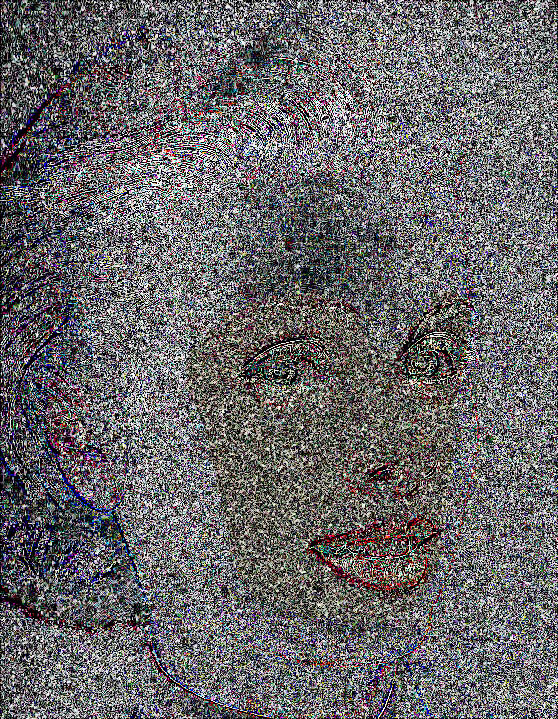}
\label{fig:f_grace2}}
\caption{Result of testing on Multi-Scale Harmonization (Figure~\ref{fig:harm4}). On the left, aligned Double JPEG compression was used. On the right, noise analysis. Notice that the suspect region is an almost exact outline of the composition mask (Figure~\ref{fig:harm3}). In the original composited image (Figure~\ref{fig:harm4}), however, it is impossible to notice this discrepancy visually.}

\end{figure}

The CFA analysis was done using the technique from~\cite{ferrara}, which tries to measure the presence of demosaicing artifacts at a local level onto the image. It is a fragile trace, as almost any modification can destroy it. This means that it can be really useful to decide if an image is not native (FD1) or where it has been tampered (FD2) when the image is expected to be pristine. Since the modification history of our images was unknown, some test cases were inconclusive because the demosaicing pattern had already been disrupted.

\begin{figure*}[ht!]
\centering
\subfloat[]{\includegraphics[width=0.2339\textwidth]{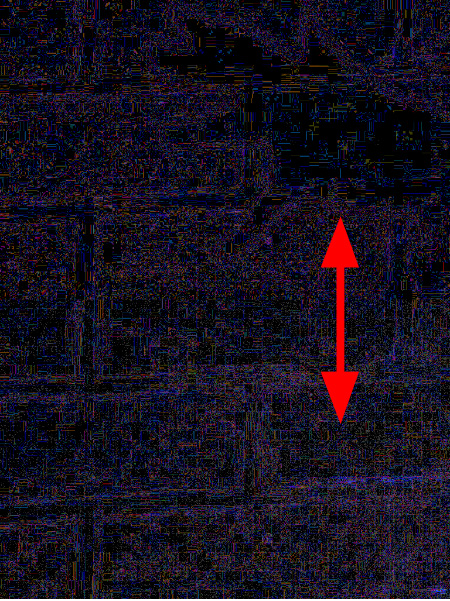}
\label{fig:f_unshadow}}
\subfloat[]{\includegraphics[width=0.416\textwidth]{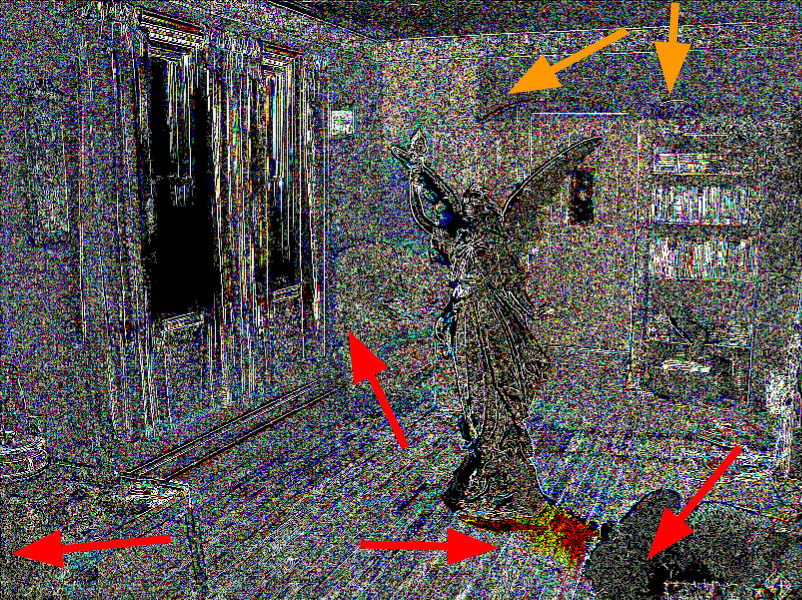}
\label{fig:f_insert2}}
\subfloat[]{\includegraphics[width=0.2339\textwidth]{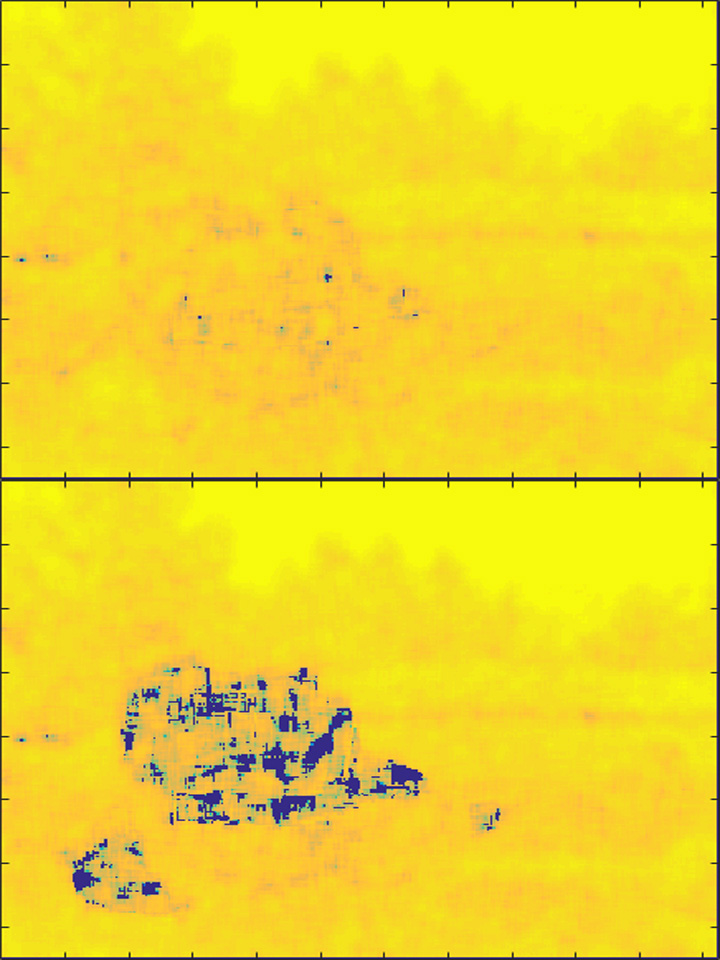}
\label{fig:f_localizedrecoloring}}
\\
\subfloat[]{\includegraphics[width=0.208\textwidth]{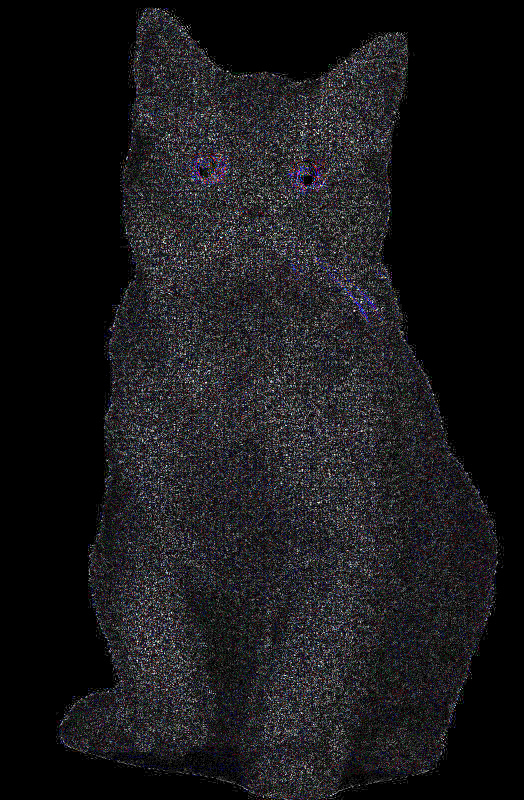}
\label{fig:f_morph1}}
\subfloat[]{\includegraphics[width=0.208\textwidth]{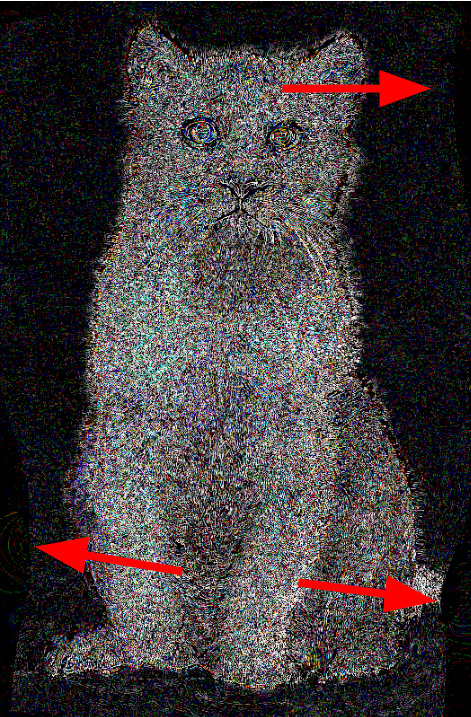}
\label{fig:f_morph2}}
\subfloat[]{\includegraphics[width=0.208\textwidth]{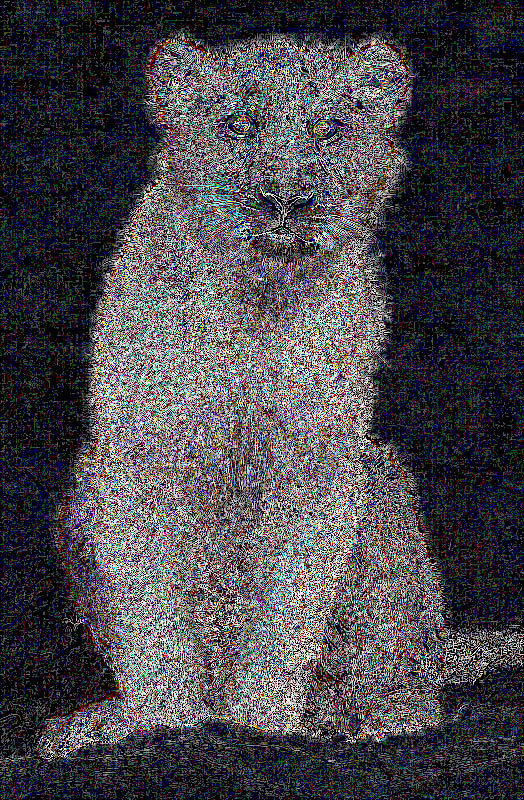}
\label{fig:f_morph3}}
\subfloat[]{\includegraphics[width=0.256\textwidth]{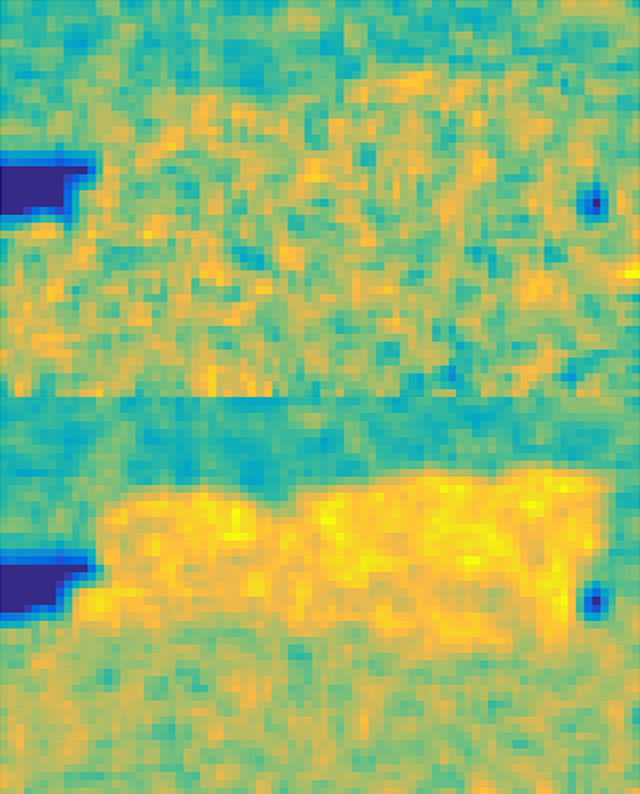}
\label{fig:f_reillum}}

\caption{Results of analyzing different traces for the images on Figure~\ref{fig:compositingexamples1}. a) ELA of Soft Shadow Removal. In this case, it is not possible to identify any irregularity in the composited image. b) Noise analysis of object insertion. The first identifiable irregularity is that the noise pattern for the shadow cast by the synthetic object greatly differs from other shadowed regions in the image (red arrows). The indirect illumination estimated after the scene's light interactions with the object appear as salient planes in the noise map (orange arrows). c) PRNU analysis of localized recoloring. The more yellow, higher is the correlation between the region and the cameras sensor pattern noise. On the first image, there are some false positives thorough the image caused by high frequency areas. On the recolored image, the probability map shifts completely to the altered region. d-f) Noise analysis of image morphing. The morphing process creates distinct warping artifacts on the noise pattern. g) Double JPEG compression analysis of reillumination. The more yellow, higher the probability that the region has undergone double JPEG compression. While the top image shows a very noisy pattern, in the bottom image the uniform interpretation of a salient portion suggest that different compression traces (single and double) are present in the image. }
\label{fig:f_compositingexamples}
\end{figure*}

Unfortunately no information about source device were available for any of the provided images. This limits the the type of analysis that can be done, for instance using PRNU-based analysis~\cite{FridrichPRNU} to localize forgeries. Since the PRNU is a prominent trace used in the literature, we designed an experiment to test it. First, we captured a set of uncompressed images with a Canon 5D to be able to estimate the noise pattern. We were able to create compositions with the images using Alpha Matting~\cite{Gastal2010} and Edge-Aware Filtering~\cite{GastalOliveira2015}, both of which the implementation was easily available. In both cases, the PRNU analysis was able to identify the forgeries. In Fig.\ref{fig:f_localizedrecoloring} we report an example to show its effectiveness on localized recoloring. The blue region, representing a portion in which the pattern noise is uncorrelated with the one estimated by the camera, clearly localize the tampering.

Error Level Analysis (ELA) estimates the local compression error for each JPEG block. This is a very low-level trace that is common on forensics analysis tools, but requires a great deal of interpretation and can be misleading~\cite{Wang2011}. It is, however, an interesting analysis tool for our testing purposes. It allows us to compare differences in compression for images before and after the composition techniques were applied, rather than measuring a specific trace. In our tests (Table~\ref{tab:ImageTesting}), several techniques obtained a ``plausible" result, which indicates that compression traces could be used to detect them. In Fig.~\ref{fig:f_unshadow} ELA is not able to identify any irregularity in the composed image while, in both~\ref{fig:f_insert2} and~\ref{fig:f_morph2} the forgery introduces distinct noise artefacts.

To evaluate double JPEG compression the technique from~\cite{bianchi2012} was used. The results are more straight-forward to analyze than ELA, but its scope is more limited. It is specially useful if a foreign region is inserted in an image (object transfer/insertion). Reillumination (Fig.~\ref{fig:f_reillum}) and seamless cloning (Fig.~\ref{fig:f_grace}) were easily identified in our tests, and our evidence indicates that the same could apply to Alpha Matting.

Noise analysis is very simple and subjective, but can be very powerful. A median filter is applied on the image, and the result is subtracted from the original one. The result is then scaled according to the local luminance, and the noise pattern is obtained. In essence, we are observing a combination of sensor noise, illumination noise, compression noise, and high frequencies. In natural images, the noise tends to behave similarly in almost all of the image. When a region has a distinct noise pattern (Figs.~\ref{fig:f_insert2} and ~\ref{fig:f_grace2}), or strange artifacts appear (Fig.~\ref{fig:f_morph2}), it can be a strong indicator of forgery. 

The majority of tested composition types had at least one identifiable trace. This means that forgeries done using these techniques would probably be identified with a careful analysis. In some cases, there is also the possibility of developing a specific technique to detect them. 

\section{Conclusion}
\label{sec:conclusion}
In this work we surveyed both the fields of Image Composition and Digital Image Forensics, crossing them directly. A new classification scheme for techniques in both fields, along with a forgery detection scale were presented to help organize the discussion. This scale provides a more sophisticated way to compare the existing works on the literature, and assess the capabilities of forensic techniques. To understand the forensics aspect of composition techniques, their inner workings were studied and tests were performed for a wide variety of image effects. As a result, we uncovered that current state-of-the-art forensics has all the basic tools it needs to be able to detect most forgeries, provided that they are properly tuned for each specific case, and maybe even combined. 

This work is an overview of both fields, and can be used either as an introductory reading, or as summary of the current challenges. In this sense, we provide groundwork for the development of novel forensic techniques against image composition. A natural extension for this work would be to increase the number of techniques surveyed and tested, considering other traces and forensic approaches. Since both fields are in an ``eternal arms race" the list of available works to be compared will keep increasing each year. We have shown, however, that it is not trivial to fool a careful and thorough forensic analysis.

\section*{Acknowledgment}
The authors are partially supported by GNSAGA of INdAM, CAPES and CNPQ (grants 200960/2015-6/SWE and 306196/2014-0). We also thank all the authors that provided us with images and material to perform our tests.

\bibliographystyle{IEEEtran}
\bibliography{SurveyForensicsComposition}

\end{document}